\documentclass[10pt,twocolumn,letterpaper]{article}

\usepackage{iccv}
\usepackage{times}
\usepackage{epsfig}
\usepackage{graphicx}
\usepackage{amsmath}
\usepackage{amssymb}
\usepackage{caption}

\usepackage{times}
\usepackage{epsfig}
\usepackage{graphicx}
\usepackage{amsmath}
\usepackage{amssymb}
\usepackage{bbm}
\usepackage{algorithm}
\usepackage{algpseudocode}
\usepackage{xcolor}
\usepackage{enumerate}
\usepackage{enumitem}
\usepackage{booktabs}
\usepackage{pifont}

\usepackage[pagebackref=true,breaklinks=true,letterpaper=true,colorlinks,bookmarks=false]{hyperref}

\iccvfinalcopy % *** Uncomment this line for the final submission

\def\iccvPaperID{****} % *** Enter the ICCV Paper ID here

\ificcvfinal\fi

\begin{document}

%%%%%%%%% TITLE
\title{Multi-Object Tracking with Hallucinated and Unlabeled Videos}

\author{Daniel McKee$^{1,2}$\hspace{8mm}Bing Shuai$^2$\hspace{8mm}Andrew Berneshawi$^2$\hspace{8mm}Manchen Wang$^2$ \\
Davide Modolo$^2$\hspace{8mm}Svetlana Lazebnik$^1$\hspace{8mm}Joseph Tighe$^2$ \vspace{0.3em}\\
$^1$University of Illinois at Urbana-Champaign\hspace{12mm}$^2$Amazon Web Services \\
}

\maketitle
\ificcvfinal\thispagestyle{empty}\fi

\begin{abstract}

In this paper, we explore learning end-to-end deep neural trackers without tracking annotations. This is important as large-scale training data is essential for training deep neural trackers while tracking annotations are expensive to acquire. In place of tracking annotations, we first hallucinate videos from images with bounding box annotations using zoom-in/out motion transformations to obtain free tracking labels. We add video simulation augmentations to create a diverse tracking dataset, albeit with simple motion. Next, to tackle harder tracking cases, we mine hard examples across an unlabeled pool of real videos with a tracker trained on our hallucinated video data. For hard example mining, we propose an optimization-based connecting process to first identify and then rectify hard examples from the pool of unlabeled videos. Finally, we train our tracker jointly on hallucinated data and mined hard video examples.  Our weakly supervised tracker achieves state-of-the-art performance on the MOT17 and TAO-person datasets. On MOT17, we further demonstrate that the combination of our self-generated data and the existing manually-annotated data leads to additional improvements.

\end{abstract}

%%%%%%%%% BODY TEXT
\section{Introduction}

Recent progress on deep learning based trackers \cite{bergmann2019tracking, meinhardt2021trackformer, shuai2021siammot, zhou2020tracking} has elevated the performance of online multiple-object tracking (MOT) to a level that is comparable to offline trackers \cite{berclaz2011multiple,braso2020learning,tang2017multiple,xiang2015learning,xu2019spatial,zhang2008global}.
This new family of trackers includes two fundamental components: detection of objects in a frame and association of detected objects across frames into tracks (often by estimating object motion). 
Training these trackers requires both a large number of bounding box annotations for the detection component and a large number of instance-level correspondence annotations for the motion model.
Unfortunately, large-scale instance correspondence annotations are extremely expensive to obtain. As a consequence, the current MOT datasets are usually restricted to a limited number of videos (e.g.~7 training videos for MOT17 \cite{MOT16}). While these trackers can be trained with a large set of image-based bounding box annotations, the amount of tracking annotations available for training is relatively small. This raises the question: \textit{What if we had large-scale tracking datasets available for training?}

We conjecture that a large-scale dataset is essential to training robust deep neural models for tracking, as is common in training deep networks for other vision tasks \cite{deng2009imagenet,he2016deep,lin2014microsoft,ren2015faster}. Thus, in this work, we explore ways to auto-annotate large sets of unlabeled videos such that we can leverage them for training a deep neural tracker. Specifically, we explore auto-annotation in a weakly supervised setting, where we have bounding box annotations for images and a large pool of unlabeled videos.

\begin{figure}
    \centering
    \includegraphics[width=0.5\textwidth]{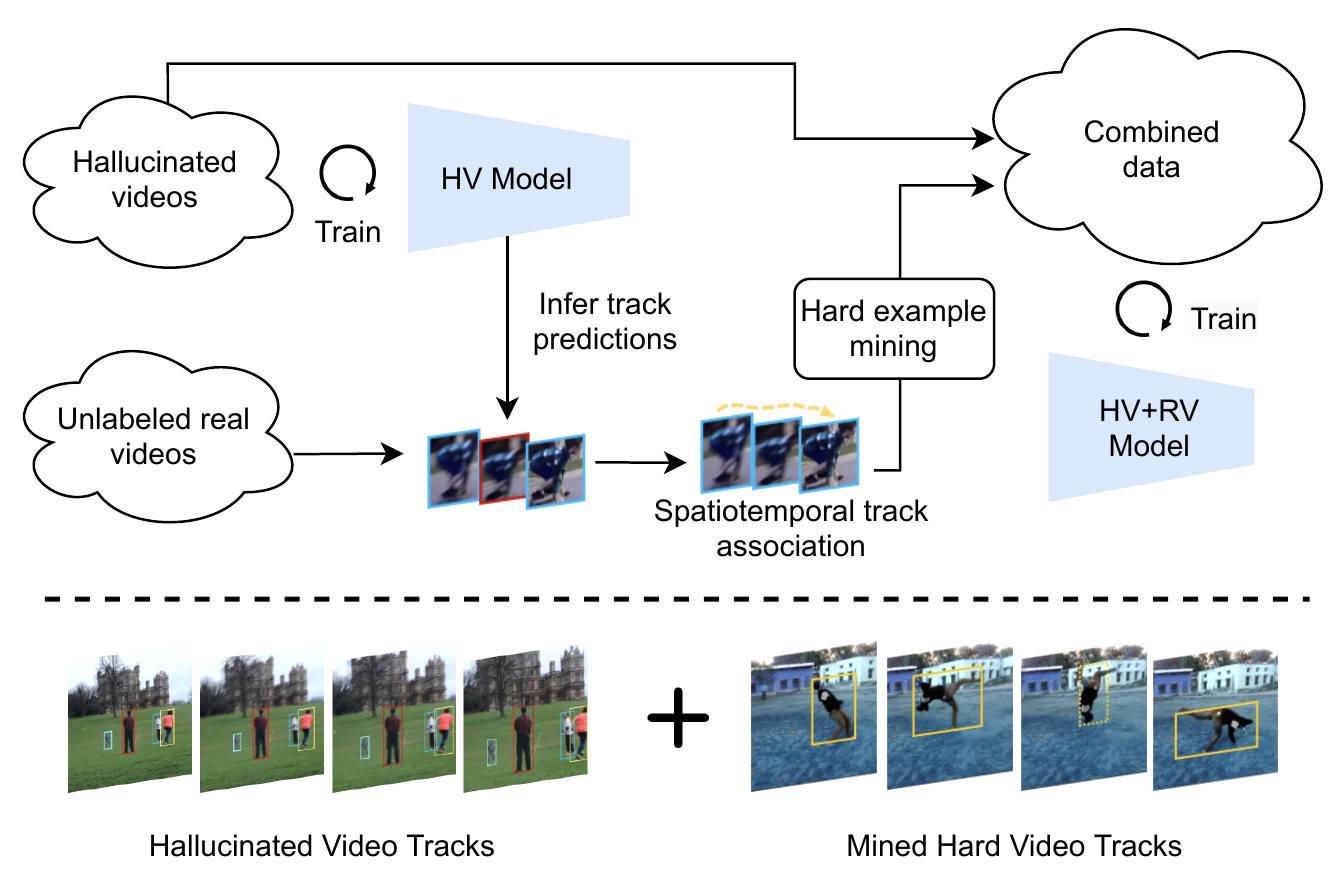} 
    \caption{\small We first generate a dataset of hallucinated videos (HV) from images on which we train an HV tracking model. Next, we mine a set of hard examples from unlabeled real videos (RV). We train the tracking model jointly on the HV and RV datasets. We hallucinate a short video from a single image with a recurrent zoom-in/out motion transformation as well as with other random motion effect transformations. Hard examples refer to videos where the HV-trained tracker fails, such as a video in which a person's pose deforms dramatically. 
    }
    \label{fig:teaser}
\end{figure}

Starting with a single image with bounding box annotations, we hallucinate a short video with bounding box instance correspondences.  To do this, we apply a zoom-in/out transformation to the image and bounding boxes respectively. Our zoom-in/out transformation is similar to the scaling and translation operation used in previous tracker training \cite{held2016learning, zhou2020tracking}, but we extend this notion with advanced motion blur, lighting effect changes, and other video artifact simulations. We train a deep neural tracker on this set of hallucinated video data (HV) and show that it already performs competitively with the state of the art.

While useful for training, our hallucinated videos still lack visual characteristics of real videos which are hard to simulate.
Therefore, we mine on a set of unlabeled videos for hard examples where the HV tracker fails. For example, the predicted tracking trajectories are often broken when object instances are occluded, or their poses suddenly deform. We identify these hard examples by looking for tracklets that are potentially broken from a single object instance's trajectory. To this end, we optimize a video-level cost function that takes into account the spatial and temporal coherence between tracklets. 
High-confidence tracklets that are compatible with each other will be merged, generating rectified pseudo labels for the identified hard examples. We show that adding this set of rectified hard examples (RV) to our existing HV data is
important for training our weakly supervised deep neural tracker, as these examples guide the model to track through challenging scenarios in real videos.

We study the effectiveness of our mining method on the multi-person tracking task. To this end, we generate hallucinated videos from person detection image datasets: COCO \cite{lin2014microsoft} and CrowdHuman \cite{shao2018crowdhuman}, and we mine hard examples from the Kinetics \cite{kay2017kinetics} person-centric video dataset.  By testing our model on two well-studied datasets, MOT17 \cite{MOT16} and TAO-Person \cite{dave2020tao}, we make three key findings: 1) our weakly-trained tracker achieves state-of-the-art results on both the MOT17 \cite{MOT16} and TAO-person \cite{dave2020tao} datasets; 2) on the MOT17 \cite{MOT16} dataset, the tracker trained with large-scale self-generated data and annotations outperforms its counterpart that is trained on the provided MOT17 tracking annotation and importantly 3) adding our self-generated annotations to the MOT17 training dataset significantly improves the tracker's performance.  These results clearly demonstrate the value of our self-generated data and annotations even when a small amount of manual annotations are available. Moreover, we conduct thorough ablation experiments to investigate the optimal settings for hard example mining as well as how to effectively train a weakly supervised deep tracker with self-generated tracking annotations.

To summarize, we propose a deep neural tracker training framework to train from only detection annotations and unlabeled videos. The framework relies on two data-generation components: video hallucination and unlabeled video track mining.
Our hallucinated video technique builds on data generation in previous works by introducing more advanced video appearance simulation. 
Our mining process differs from previous MOT video mining approaches in that it:
1) performs a tracklet matching optimization based on spatiotemporal coherence rather than matching detections solely on spatial coherence, and 2) acts as a hard example mining process by identifying and rectifying spatiotemporal discontinuities between tracklets. 
Our experimental results validate each of these contributions and show state-of-the-art results on the MOT17 and TAO-Person datasets.

\section{Related Work}

\paragraph{Multi-Object Tracking.}
The multi-object tracking (MOT) task involves detecting objects of interest in every frame and then temporally linking them to form trajectories. Most recent MOT works fall into the ``tracking-by-detection" paradigm \cite{berclaz2011multiple,bewley2016simple,tang2017multiple,wojke2017simple,xiang2015learning,xu2019spatial,zhang2008global} which poses MOT as a data association problem. There are two different ways to to tackle this formulation. One is an ``offline tracker" approach, in which a large offline graph is first built \cite{andriyenko2011multi, berclaz2006robust, tang2017multiple, xiang2015learning, xu2019spatial} that encodes the visual similarity and geometric consistency between detected bounding boxes. Next, the trajectories are derived by optimizing the graph-induced cost function. Although these offline trackers can output trajectories with high temporal consistency, they are usually inefficient and less applicable to real-time tracking. The second approach is an ``online tracker" which performs data association on the fly \cite{bergmann2019tracking, bewley2016simple, meinhardt2021trackformer, shuai2021siammot, wojke2017simple, zhou2020tracking}. 
Recently, deep neural network based multi-object trackers \cite{bergmann2019tracking, meinhardt2021trackformer, shuai2021siammot, sun2020transtrack, zhou2020tracking} have pushed the online tracking performance to a level that is comparable to offline tracking.

In general, existing deep neural trackers consist of two key components: 1) a detector to detect new object instances and 2) a motion model to estimate the instance's movement between frames such that the detected instances can be linked across time. For example, Bergmann et al. \cite{bergmann2019tracking} reuse the second stage of Faster-RCNN \cite{ren2015faster} -- the detection branch -- as the motion model; Zhou et al. \cite{zhou2020tracking} estimate the object instance's motion by feeding the model with location maps at the previous frame together with feature maps from both previous and current frames. Shuai et al. \cite{shuai2021siammot} leverage a single object tracker to predict the instance's motion. Both Meinhardt et al. \cite{meinhardt2021trackformer} and Sun et al. \cite{sun2020transtrack} use attention-based transformers to model the instance's movement across frames. 
All of these trackers are trained with manual tracking annotations.
In this work, we explore training deep neural trackers in a weakly supervised setting by using bounding box annotations from images and a pool of unlabeled videos.

\begin{figure}
    \centering
    \includegraphics[width=0.5\textwidth]{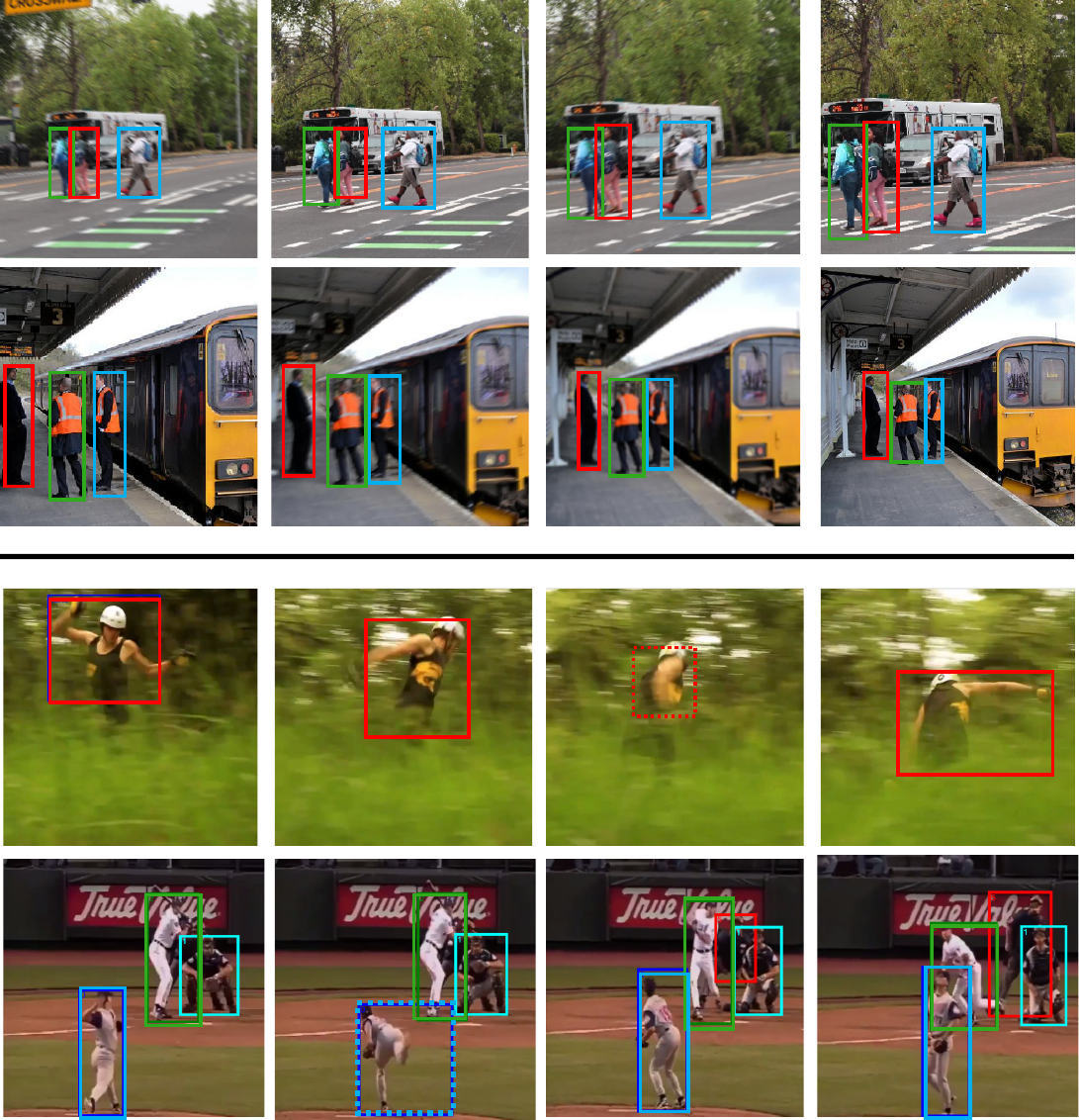}
    \caption{\small We show examples from our self-generated training data consisting of hallucinated videos (top) and mined real videos (bottom) with their corresponding pseudo-label tracks. For the real video examples we visualize pseudo-labels of mined hard examples with broken tracks (dotted box) that were identified and corrected by our track rectification process. }
    \label{fig:hv_rv}
\end{figure}

\vspace{0.5em}
\noindent\textbf{Self-training}
Self-training has long been a well-known semi-supervised learning technique \cite{scudder1965probability}. Recently, self-training has been successfully applied to improve a number of modern deep learning models on vision tasks such as image classification \cite{xie2020self,zou2019confidence}, object detection \cite{li2020improving,zoph2020rethinking}, and segmentation \cite{zhu2020improving,zoph2020rethinking,zou2018unsupervised,zou2019confidence} as well as tasks in NLP \cite{he2019revisiting,kahn2020self} and speech recognition \cite{kahn2020self}. Since we must address the gap between image and video frames, our method is especially related to self-training for domain adaptation \cite{wang2018deep,zou2018unsupervised}. However, our work differs in that self-training is also ``across-task" since we start with object detection annotations and use these to obtain tracking annotations in video. 

\vspace{0.5em}
\noindent\textbf{Video Hallucination}
Hallucinating videos from a single image has been explored in different video-based tasks. Gao et al. \cite{gao2018im2flow} propose to hallucinate a flow map from a single image, which is used to improve action recognition performance. Kanazawa et al. \cite{kanazawa2019learning} learn to hallucinate temporal context from a single image, which is used for 3D human pose estimation. Li et al. \cite{li2018flow}  propose a generative model to hallucinate an RGB image from a flow map. Held et al. \cite{held2016learning} and Zhou et al. \cite{zhou2020tracking} adopt scaling and translation operations to simulate object's geometric movement across frames. In this work, we hallucinate person's trajectories in a simulated video with different motion effects. Next, we use those videos and pseudo labels to train a deep neural tracker. The closest hallucination techniques to ours appear in Held et al. \cite{held2016learning} and Zhou et al. \cite{zhou2020tracking}, which use a special case of our video hallucination where only camera zoom-in motion is applied and only two frames are generated per video sequence.

\vspace{0.5em}
\noindent\textbf{Mining Tracks from Unlabeled Video}
While there are a variety of previous works that have used tracking through unlabeled video to improve model training, most works have used mined tracks to improve object detection performance \cite{jin2018unsupervised,misra2015watch,DBLP:conf/icra/OsepVLBL19,roychowdhury2019automatic}. Work focused on improving MOT through mining on unlabeled data is very limited. The most closely related work to ours is MOTSNet \cite{porzi2020learning} which proposes to generate a dataset automatically on unlabeled video by predicting image-level segmentation. They link these segmentations into a linear assignment process based on temporal coherence between masks to form annotations.

A key difference between this work and ours is that the track generation process for MOTSNet is based only on spatial coherence between object masks in adjoining frames. Thus, the mined tracklets will be broken in frames where the detection or segmentation model fails. In contrast, we propose a spatiotemporal connecting process linking tracklets over short-term failures, which is capable of identifying and then rectifying brief discontinuities in detection outputs.  Moreover, these rectified tracklets are used as hard examples, which play a pivotal role in training a better tracking model.

\section{Tracking Architecture: DTracker}
\label{sec:dnt}

In this paper, we adopt a tracker architecture similar to the implicit model from SiamMOT \cite{shuai2021siammot}, which achieves state-of-the-art results on the MOT17 dataset. Hereafter, we refer to this specific deep neural tracker as \textbf{DTracker}.

DTracker is a two-stage deep neural tracker that is based on two-stage Faster-RCNN \cite{ren2015faster}, so it includes three major functional sub-networks: region proposal network (RPN), detection branch, and track branch (motion model). Specifically, DTracker uses a single-object tracker -- GOTURN \cite{held2016learning} -- to model the object instance's motion movement between two frames. DTracker is trained with the loss $\ell = \ell_{rpn} + \ell_{detect} + \ell_{track}$, where $\ell_{rpn}$, $\ell_{detect}$ are standard losses related to RPN and detection branch in Faster-RCNN \cite{ren2015faster}, and $\ell_{track}$ is the motion model training loss.

In order to train DTracker, an image pair is sampled from a short video sequence (e.g. one second long). Formally, $B_t$ and $B_{t+\delta}$ denote bounding boxes that correspond to the same object instance at times $t$ and $t+\delta$ respectively.
In order to track $B_t$  from $t$ to $t+\delta$, the motion model predicts $(\hat{v}, \hat{m})$ where $\hat{v}$ is a visibility score indicating whether or not the object instance is still visible at $t+\delta$ and $\hat{m}$ is the estimated motion offset for the object from $t$ to  $t+\delta$.  The motion model is trained with the following loss:

\begin{equation} \label{eq:train_loss}
    \ell_{track} = \ell_{cls}(\hat{v}, v^{gt}) +  \mathbbm{1}[v^{gt}] \ell_{reg}(\hat{m}, m^{gt})
\end{equation} 
in which $\ell_{cls}$ is the binary cross entropy loss; $\ell_{reg}$ is a smooth $\ell_1$ regression loss; $v^{gt} = 1$ if $B_{t+\delta} \neq \emptyset$, otherwise $v^{gt} = 0$; and $m^{gt}$ is the ground truth motion representation that is derived from $B_t$ and $B_{t+\delta}$ in the same way as regression targets in Faster-RCNN \cite{ren2015faster}. 
We refer the readers to  \cite{shuai2021siammot} for a more detailed description.

\section{Weakly supervised DTracker}

A large corpus of labeled tracking annotations is required to train a generalized DTracker. To be specific, bounding box annotations are needed for training the detection model in DTracker, and instance correspondence annotations within videos are needed to train the motion model. However, acquiring large-scale and diverse video-level instance correspondence annotations is expensive. Consequently, current multi-object tracking (MOT) datasets are limited to very few videos. In practice, the detection model in DTracker is trained with large-scale image datasets, while the motion model is trained with a limited video dataset.  Therefore, the feature backbone of DTracker is heavily tuned for the detection task, and the motion model struggles to generalize in different scenes or motion patterns.

Towards the goal of training a generalized tracker without annotated instance correspondences, we adopt a weakly supervised setting.
Specifically, the input to our weakly supervised DTracker training is a set of annotated images with bounding boxes for objects of interest and a pool of unlabeled videos.
In Sec. \ref{sec:hv}, we first introduce hallucinated videos and trajectories that are used for training an initial DTracker. Then we leverage this DTracker to auto-annotate and mine for hard examples from a large corpus of unlabeled videos, as described in Sec. \ref{sec:rv}. We demonstrate that these two data sources are complementary and both substantially contribute to improving our DTracker performance.

\subsection{Video Hallucination From Images} 
\label{sec:hv}

To hallucinate short annotated videos, we simulate a zoom-in/out visual effect on images with cropping and scaling operations ($\operatorname{crop\_and\_scale}$). Specifically, for an image $\mathbf{I}_1$,  we generate a $T$ frame video sequence  $[\mathbf{I}_1, \ldots, \mathbf{I}_t, \ldots, \mathbf{I}_{T}]$, in which $\mathbf{I}_t = \operatorname{crop\_and\_scale}(\mathbf{I}_{t-1}, r*(t-1))$ for $t \geq 2$. Here, $r (\geq \frac{0.9}{T-1})$  denotes the relative size of image crop w.r.t. that of the original image $\mathbf{I}_1$. 
This zoom-in/out transformation is similar to the scaling and translation operation used in \cite{held2016learning, zhou2020tracking}. To simulate more realistic appearance in videos, we inject visual effects such as motion blur, color/lighting variations, and compression artifacts during video hallucination. 

We generate tracking annotations for the hallucinated videos by applying the same geometric zoom-in/out transformation to bounding boxes. 

As shown in Figure \ref{fig:hv_rv}, the zoom-in/out motion effects mimic the geometric and scale changes of object instances that result from natural camera and object motion. Such generated training examples can be used to learn a motion model robust to fast camera or object motion as well as common video appearance artifacts. 
As each image represents a distinct scene, the large quantity of hallucinated videos offers a great diversity of scenes, far outnumbering the existing tracking datasets.

\subsection{Hard Example Mining from Unlabeled Videos}
\label{sec:rv}

Our hallucinated videos allow us to train a DTracker with no video supervision but they do not fully capture all forms of motion found in real video. 
First, all object instances go through the same transformation, so there are no instances of objects with different trajectories crossing in front of each other. Second, our hallucinated videos have no deformations of object instances (i.e. a person changing pose). And third, there are no cases of changing occlusion in our hallucinated videos. To overcome these data limitations, we look to real video data.

To generate this real video training data, we start with a pool of unlabeled videos and run our hallucinated video trained DTracker over these videos. Our DTracker produces high confidence short tracks (tracklets) but fails to track objects through the challenging video specific scenarios such as occlusion or articulation of the object, as demonstrated in Figure \ref{fig:hv_rv}. Thus, we target our hard example mining method to these gaps in the tracking output. By doing so we are able to mine examples from our unlabeled set of videos that complement the hallucinated video training dataset.

Inspired by other works which mine hard examples or rectify pseudo-labels to improve self-training for object detection \cite{jin2018unsupervised,li2020improving,roychowdhury2019automatic,zou2019confidence}, we propose a mining process that first locates hard tracking examples and then rectifies them to output pseudo labels. As Figure \ref{fig:hv_rv} shows, the hard examples usually refer to tracklets that are broken from an otherwise complete trajectory.
Those broken tracklets that belong to the same object instance are usually spatially and temporally coherent, especially in videos with the less-crowded scenes. Therefore, we propose the pairwise matching cost between two tracklets as follows:
\begin{equation} \label{eq:objective}
    c(i, j) = tIoU(\tau_i, \tau_j) + (1 - \frac{t(\tau_i, \tau_j)}{\gamma}) + \chi(\tau_i, \tau_j)     
\end{equation}
where $tIoU(\tau_i, \tau_j)$ computes the spatial IoU between the end of track $\tau_i$ and the beginning of track $\tau_j$ while  $t(\tau_i, \tau_j)$ computes the time gap between the two tracks. $\chi(\tau_i, \tau_j)$ is a characteristic function to impose hard constraints on matching between tracks, defined as: 
\vspace{-8pt}
\begin{equation}
\chi(\tau_i, \tau_j) = 
     \begin{cases}
       -\infty &\quad\text{if} \quad tIoU(\tau_i, \tau_j) < \mu \\
        &\quad\quad\text{or} \quad t(\tau_i, \tau_j) > \gamma\\
        &\quad\quad \text{or} \quad i \leq j \\
       0 &\quad\text{otherwise} \\ 
     \end{cases}
\label{eq:cost_function}
\end{equation}

In order to identify and rectify the broken tracklets, we propose an optimization process that connects the broken tracklets that belong to the same object instance. We define the optimization problem as follows:
\begin{equation}
\begin{aligned}
    \max_{x} \quad \sum_{i,j} c(i, j) x_{ij} \\
    \textrm{s.t.} \sum_i x_{i,j}  \leq 1 \quad \forall  j \\
    \sum_j x_{i,j} \leq 1 \quad  \forall  i  \\
    x_{ij} \in \{0,1\} \quad \forall i, j \\
\end{aligned}
\end{equation}
in which we use a binary variable $x_{ij}$ to denote the connectivity between two tracks $(\tau_i, \tau_j)$. A value of 1 corresponds to joining two tracks $(\tau_i, \tau_j)$. At each step, we impose constraints that a track may only be joined to one later track and one earlier track. This prevents cases where a single track might be split into multiple separate overlapping tracks. 
We optimize this process in an iterative manner:  first running the optimization process to obtain a set of tracks to join, then updating all tracks, and repeating until convergence when no additional tracks are matched. 
$x_{ij} = 1$ indicates the identification of a hard example, which is the video sequence covering both tracklet $\tau_i$ and $\tau_j$. Connecting tracklets ($\tau_i, \tau_j$ for $x_{ij}=1$)  gives rise to the rectified pseudo labels. We show some mined hard examples in Figure \ref{fig:hv_rv}, in which broken tracklets are connected in challenging scenarios.  In this way, we generate a training dataset that is focused on the failure modes of the hallucinated video trained DTracker.

\paragraph{Noise in pseudo labels.} Our hard example mining process will inevitably have noise from the pseudo labels through either tracklet prediction errors or tracklet association errors in our optimization process. To mitigate model corruption due to this noise, we propose to regularize the model training with a joint training scheme, in which every batch includes a proportion of hard examples and the rest are hallucinated videos.  As pseudo labels for hallucinated videos are free of noise, they help to balance out corruption in model accuracy due to mined video noise.

\section{Experiments}
\subsection{Implementation details}
We use DLA-34 \cite{yu2018deep} as the feature backbone of DTracker, which adds the motion model -- GOTURN \cite{he2016deep} -- to the 2$^{nd}$ stage of Faster-RCNN \cite{ren2015faster}.

\vspace{0.5em}
\noindent\textbf{Motion model.}
Given a tracked target at time $t$ with bounding box $B_t$, we define the target's corresponding contextual regions at times $t$ and $t+\delta$ to be $C_t$ and $C_{t+\delta}$ respectively. The contextual regions share the same geometric center as that of $B_t$, but with $2 \times$ the width and height of $B_t$. In order to track the target from time $t$ to $t+\delta$, the motion model in DTracker takes the visual features of $C_t$ ($\mathbf{f}_{C}^t$) and $C_{t+\delta}$ ($\mathbf{f}_{C}^{t+\delta}$) and outputs its visibility score $\hat{v}$ and motion offset $\hat{m}$. Note that the ROIAlign operator \cite{he2017mask} is used to pool $\mathbf{f}_C^{t}$ from backbone feature maps with $C_t$. We use two fully connected layers with 512 hidden neurons to represent the motion model, and it is jointly trained with the rest of Faster-RCNN. 

\vspace{0.5em}
\noindent\textbf{Video hallucination and mining.}
We generate hallucinated videos from COCO-2017 \cite{lin2014microsoft} and CrowdHuman \cite{shao2018crowdhuman} image datasets. We only consider training images that include person bounding box annotations, which yields a total combined set of 78,832 images. The hallucinated videos are set to be $T=16$ frames in temporal length, and we use the public $img\_aug$ package to implement the visual transformation functions. We mine hard examples from the train split of the Kinetics-700 dataset \cite{carreira2019short,kay2017kinetics} that consists of around 545k videos. The Kinetics dataset is primarily used for human activity recognition, so almost all videos include at least one person. We empirically set $\mu$ and $\gamma$ (in Eq. \ref{eq:cost_function}) to be 0.1 and 0.5 during the hard example mining process.

\vspace{0.5em}
\noindent\textbf{Training.}
We stochastically sample two frames from either a hallucinated video or a short real video clip to be the training example. We train our model with batches of 16 training examples for a total of 25k iterations, unless specified.  We train with SGD using momentum 0.9 and weight decay 0.0001. The learning rate is set to 0.02 and decreased by a factor of 10 at 60\% and 80\% of its full iterations. As we mine hard examples from unlabeled videos primarily for training the motion model in DTracker, we ignore the detection losses calculated from them. During training, every video is resized to have a shorter dimension randomly sampled from (640, 720, 800, 880, 960), while its longer dimension is capped at 1500 pixels. For every training batch, we sample 50\% examples from mined real videos, and the rest from hallucinated videos.

\vspace{0.5em}
\noindent\textbf{Inference.}
We use the standard online inference based on SORT. A track (output from motion model) is continued if its visibility score is above threshold ($ vis \geq 0.3$) and it spatially matches with a detection (IOU $\geq 0.5$). A track is terminated if its visibility score is below $0.3$. Finally, a new track is spawned for an unmatched high-confidence detection ( $\geq 0.5$). 

\begin{table}[t]
\small
\centering
\begin{tabular}{l|@{\hskip 0.5em}c@{\hskip 0.5em}c@{\hskip 0.5em}c@{\hskip 0.5em}c@{\hskip 0.5em}c@{\hskip 0.5em}c}
\toprule
Method &  MOTA $\uparrow$ & IDF1 $\uparrow$  & FP $\downarrow$ & FN $\downarrow$ & IDsw $\downarrow$ \\
\midrule
Tracktor++ \cite{bergmann2019tracking}   & 53.5 & 52.3  & 12201 & 248047 & 2072 \\
DeepMOT \cite{xu2019train}   & 53.7 & 53.8 &  11731 & 247447 & 1947 \\
Tracktor++v2 \cite{bergmann2019tracking}  & 56.5 & 55.1  & \textbf{8866} & 235449 & 3763 \\
NeuralSolver \cite{braso2020learning}  & 58.8 & 61.7  & 17413 & 213594 & \textbf{1185} \\
CenterTrack \cite{zhou2020tracking}  & 61.5 & 59.6 &  14076 & 200672 & 2583 \\
DTracker & 60.7 & 58.8 & 15235 & 204373 & 2169 \\
\midrule
WDTracker & 64.7 & 61.4 & 11970 & 185403 & 2023 \\
WDTracker$^{*}$ & \textbf{65.8} & \textbf{63.5} & 13076 & \textbf{177893} & 1883 \\ 
\bottomrule
\end{tabular}
\caption{\small Comparison with state-of-the-art methods on MOT17 test set using \textbf{public detections}. All methods in the first block, including DTracker, are trained with MOT17 training dataset. WDTracker denotes weakly-supervised DTracker that is only trained with our self-generated dataset.
WDTracker$^*$ represents DTracker that is trained with the combination of MOT17 and our self-generated dataset. }
\label{mot-results}
\end{table}

\begin{table}[t]
\small
\centering
\vspace{0.5em}
\begin{tabular}{lcc}
\toprule
Model                   &  AP@0.5 &  AP@0.75  \\
\midrule
Tracktor \cite{bergmann2019tracking}   &            25.9         & -                \\
Tracktor++ \cite{bergmann2019tracking} &          36.7         & -              \\
\midrule
WDTracker                      & 40.7	& 22.0           \\
WDTracker+ReID                  & \textbf{44.8}         & \textbf{25.3}          \\
\bottomrule
\end{tabular}
\caption{\small Comparison with state-of-the-art methods on TAO-Person validation set.}
\label{tao-sota}
\end{table}

\subsection{Evaluation Datasets}
\paragraph{MOT17.} The MOT17 dataset includes 7 training and 7 test videos, each video has a duration ranging from 20 to 90 seconds. It is the de-facto standard benchmark for multi-person tracking and is used in the MOTChallenge \cite{MOTChallenge2015,MOT16}. MOT17 videos focus on tracking persons in crowded scenes captured at locations such as outdoor plazas or indoor shopping malls. Therefore, it is of great difficulty to track a person consistently due to the common occlusions in crowds.

\paragraph{TAO-Person.} The TAO dataset \cite{dave2020tao} was recently released for large-scale general object tracking. Each video is around 30 seconds in duration. In this work, we only use the TAO-Person subset as we specifically focus on person tracking. In detail, TAO-Person includes 418 train videos and 826 validation videos. However the training videos are not exhaustively annotated, therefore they are not designed for model training but primarily for parameter tuning for inference. Similar to earlier works \cite{dave2020tao}, we report evaluation on the validation split, and we only use the train set for tuning hyperparameters. Tracking people in TAO-Person videos is particularly challenging due to large person pose deformation, heterogeneous person motion, fast camera motion, and motion blur artifacts, etc.

\subsection{Evaluation metrics}
We report standard tracking metrics on both TAO-person and MOT17. Concretely, by following \cite{dave2020tao}, we report Federated Track AP \cite{dave2020tao} on the TAO-person dataset. Following the same protocol with earlier work \cite{bergmann2019tracking, meinhardt2021trackformer, MOT16, shuai2021siammot, zhou2020tracking}, we report MOTA, IDF1, and related metrics on MOT17. Note that TrackAP highlights temporal consistency of the underlying tracks, whereas MOTA usually emphasizes the frame-wise recall and precision of the per-frame bounding boxes. 

\section{Results} \label{sec:results}

\begin{figure}
    \centering
    \begin{tabular}{c}
    \includegraphics[width=0.5\textwidth]{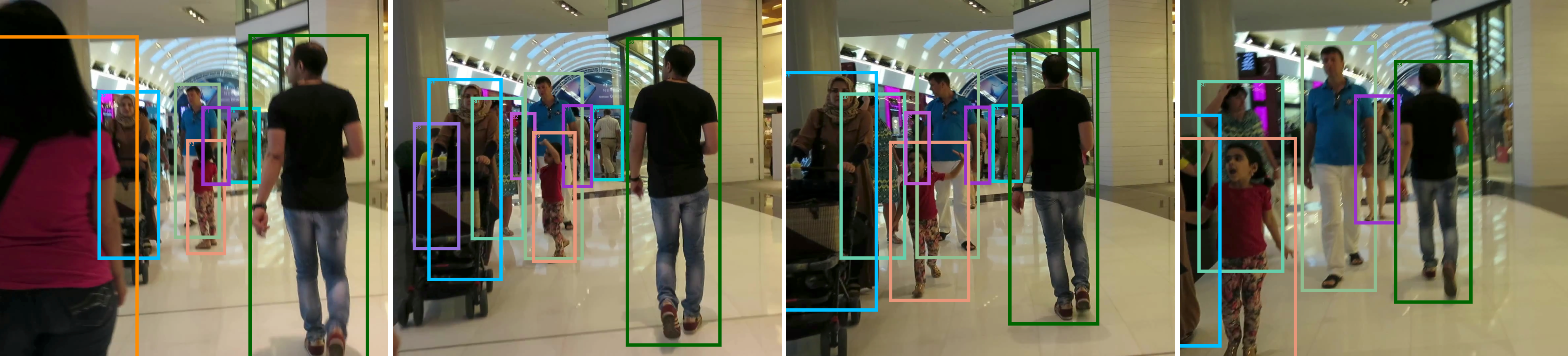} \\
    \includegraphics[width=0.5\textwidth]{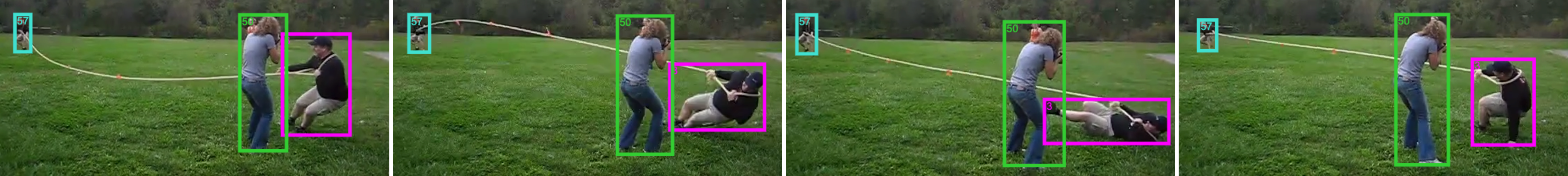} \\
    \includegraphics[width=0.5\textwidth]{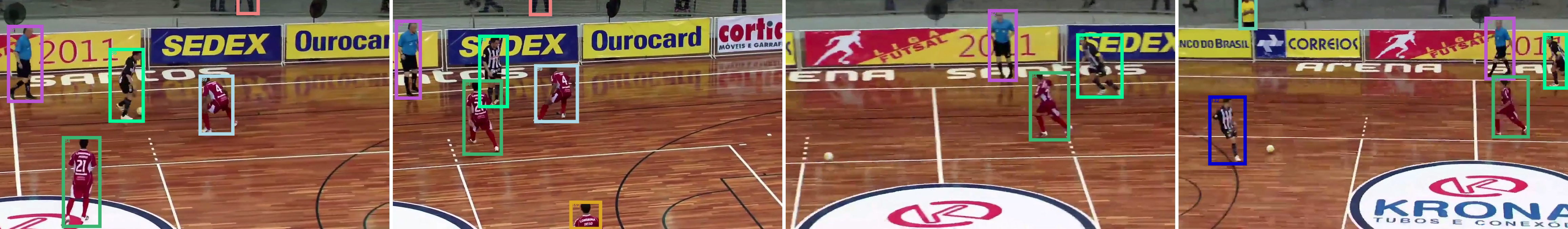} \\
    \end{tabular}
    \captionof{figure}{ \small
    The tracker trained with our hallucinated and auto-annotated videos is able to track through very challenging scenarios including occlusion in crowded scenes, large pose deformations, and fast object motion. The example in the first row is from the MOT17 test set, and remaining examples are from the TAO-person validation set. }
    \label{fig:qual-results}
\end{figure}

In this section, we present our Weakly supervised DTracker (WDTracker) on public multi-person tracking datasets including MOT17 \cite{MOT16} and TAO-person \cite{dave2020tao}.

\subsection{Results on MOT17}
We start by training a DTracker on the 7 training videos in MOT17 following common practice in other methods. 
For our WDTracker, as person bounding boxes in MOT17 are amodal, we use images from CrowdHuman dataset to generate our hallucinated videos as they include amodal bounding box annotations as well. That offers us 15,000 hallucinated videos with diverse scenes and settings compared to the limited 7 training videos available in MOT17. Our WDTracker is then trained with both hallucinated videos and hard mined self-generated annotations from real videos. We report the results on the MOT17 test set with standard metrics \cite{MOT16} including MOTA, IDF1, false positives, false negatives, and identity switches. In order to separate out the effects of detection, we generate the results with the public detections \cite{MOT16}.

In Table \ref{mot-results} we present our MOT17 trained DTracker along with our WDTracker trained on both hallucinated and real mined video data. DTracker performs on par with other state-of-the-art trackers but our WDTracker achieves state-of-the-art results (64.7 MOTA / 61.4 IDF1), which  outperforms existing best models (i.e. CenterTrack \cite{zhou2020tracking}) by a significant margin (3.2+ MOTA, 1.8+ IDF1). It's important to note that state-of-the-art models are all trained on the MOT17 dataset. This result suggests that our large-scale self-generated datasets are as valuable as small-scale manually-annotated datasets. To further validate this hypothesis, we report results for another variant of our model, WDTracker$^*$, that trains the underlying DTracker jointly on our self-generated dataset and MOT17. Interestingly, we observe another non-trivial performance boost (1.1+ MOTA and 2.3+ IDF1), which indicates that our self-generated datasets are complementary to existing tracking manual annotations.

\subsection{Results on TAO-Person}
We train our WDTracker with hallucinated videos from the COCO-17 and CrowdHuman datasets together with hard examples mined from the Kinetics dataset. We also adopt a larger feature backbone DLA-102, which is still a lighter network compared to ResNet-101 used for Tracktor++ \cite{dave2020tao}. We train the model for 50K iterations in total (2x our default training).  
As shown in Table \ref{tao-sota}, our WDTracker outperforms Tracktor and Tracktor++ by a substantial margin. We also apply the same person re-identification network and techniques as in \cite{dave2020tao} to link tracklets, our WDTracker is further improved to 44.8\% AP@IOU=0.5, which sets a new state-of-the-art performance.  These results show the clear advantage of our WDTracker as a generalizable tracker on the highly diverse and challenging examples in TAO. It's important to note that we are unable to present results of DTracker or WDTracker$^*$ analogous to those in Table \ref{mot-results}, as TAO-person dataset does not have appropriate annotations for deep model training. We show some qualitative tracking results in Figure \ref{fig:qual-results}.

\begin{table}[t]
\small
\centering
\begin{tabular}{llcc}
\toprule
Model                              &  Training data  & AP@0.5 &  AP@0.75 \\
\midrule
Tracktor                           & HV    & 28.1         & 13.3     \\
Tracktor + Flow                    & HV    & 32.5         & 16.4        \\ 
\midrule
WDTracker                        & HV$^-$   &  30.4      & 15.1 \\
WDTracker                   & HV    & 32.0         & 15.4    \\
WDTracker                   & HV+RV  & 35.4	& 17.0 \\
\bottomrule
\end{tabular}
\caption{\small \textbf{Ablation experiments on TAO-person dataset.} HV and RV represents hallucinated videos and real videos respectively, while HV$^-$ denotes hallucinated videos with only zoom-in/out motion effects. }
\label{main-results}
\end{table}

\begin{table}[t]
\small
\centering
\begin{tabular}{ccc}
\toprule
Hard sampling rate & AP @0.5 & AP @0.75 \\
\midrule
n/a   & 32.5 &	16.4 \\
0.00  & 32.4 &	16.2 \\
0.25  & 34.0 & 	16.9\\
0.50  & 35.0 & 17.1\\
0.75  & 35.4 & 17.0\\
1.00  & 33.3 & 15.6 \\
\bottomrule
\end{tabular}
\caption{\small \textbf{Hard example sampling rates.} Comparison of trackers trained with different hard examples (from RV) sampling rates evaluated on TAO-Person.  We report the result with no biased sampling in the first row. }
\label{hard-resample}
\end{table}

\section{Ablation Analysis}
\label{sec:abaltion}

We conduct our ablation analysis on the large-scale TAO-Person dataset, which covers a diverse set of tracking challenges including fast camera motion, dramatic human pose deformation, motion blur, severe occlusion in crowds, etc. These diverse and challenging motion patterns are not as represented in the MOT17 dataset, making TAO more appealing to validate the usefulness of hard example mining. 

As a first step, we train a DLA-34 based Tracktor on hallucinated videos (HV) generated from COCO-17 and CrowdHuman datasets. As listed in Table \ref{main-results}, our implemented Tracktor achieves 28.1\% Track AP @ IOU=0.5, which significantly outperforms the one reported in \cite{dave2020tao} (i.e. 25.9\% in Table \ref{tao-sota}).  Furthermore, we adopt a competitive baseline by estimating the person's movement with optical flow. In detail, we estimate the person's motion offset between two frames by taking the median motion vector of all constituent pixels within the person's bounding boxes.  To extract optical flow, we use PWC-Net that is trained on FlyingChairs \cite{IMKDB17}, FlyingThings3 \cite{IMKDB17}, Sintel \cite{Butler:ECCV:2012} and KITTI \cite{Menze2015CVPR}. As expected, adding flow to Tracktor improves Track AP @IOU=0.5 by a substantial 4.4\%.

\subsection{Hard example mining}
We train our WDTracker, which is effectively Tracktor plus a motion model, on our hallucinated video (HV) dataset. As demonstrated in Table \ref{main-results}, it achieves 32.0 \% AP@IOU=0.5, which is 3.9\% higher than Tracktor. This result shows that the motion model learned from hallucinated videos is able to generalize to real videos and improve tracking. We expect that the motion model is capable of following the person's trajectories when the subject is moving fast, a condition in which Tracktor fails.  Moreover, it performs comparably to our strong baseline -- Tracktor + Flow, which suggests that the learned motion model is as good as a fully supervised flow model to predict the person's movement. We also train our WDTracker on a simpler set of hallucinated videos (HV$^-$) that don't have our advanced motion effect simulation (i.e. motion blur, light changes, video compression artifacts). The model under-performs the one trained on HV by a clear 1.6\% AP@IOU=0.5 margin, which shows that advanced video augmentation is an important aspect of training a high performing DTracker.

\begin{figure}
    \centering
        \includegraphics[width=0.5\textwidth]{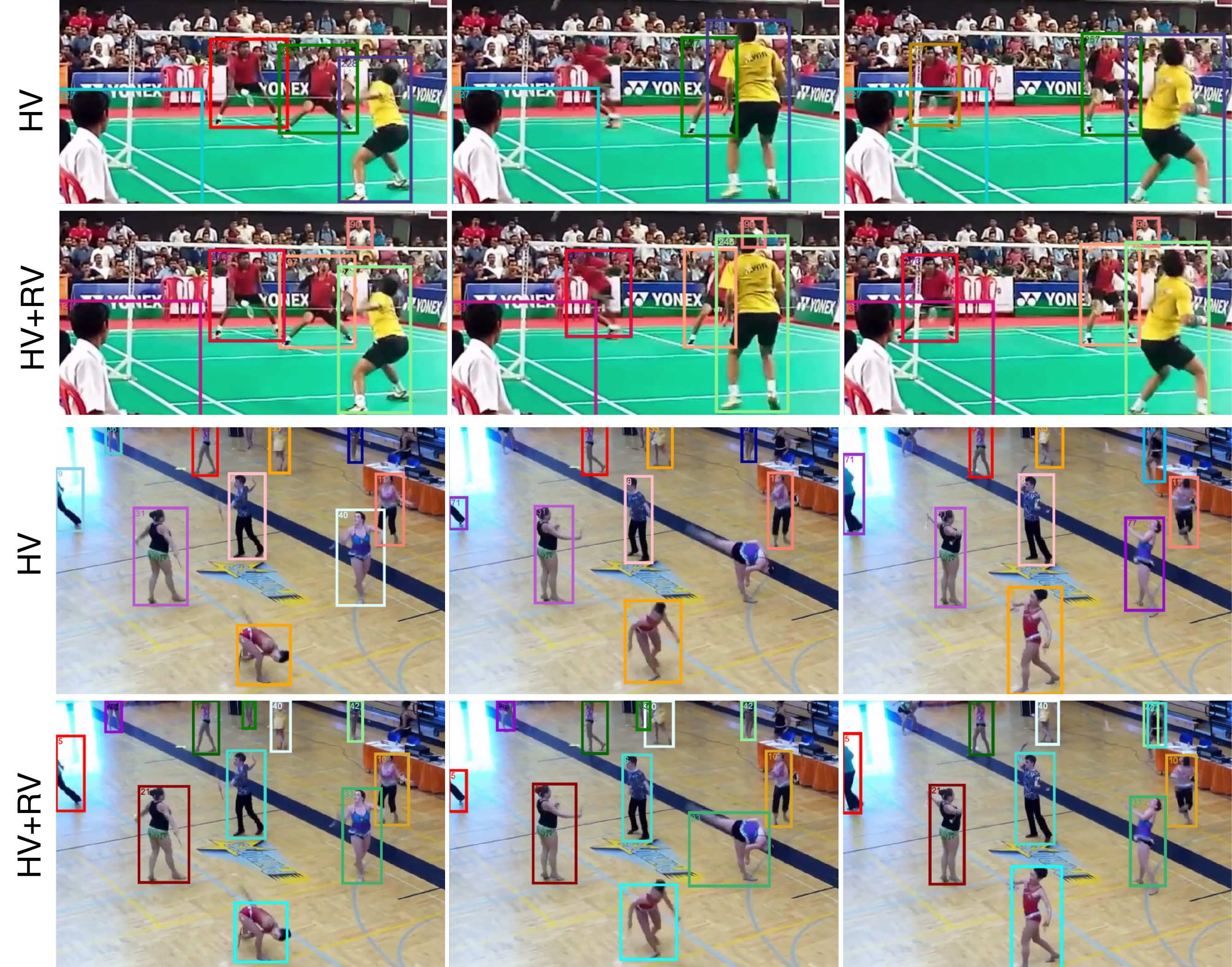} \\
    \captionof{figure}{\small In both qualitative examples from TAO-Person, the WDTracker trained only on hallucinated videos (top) fails in a case of significant pose change or movement while the model trained with hard examples from our RV dataset is robust to these challenges. In the upper half, the track is broken in the HV model for the man running across the court. In the bottom half, a track is broken during significant pose changes by the dancer on the right.}
    \label{fig:comp-results}
\end{figure}

Furthermore, we re-train WDTracker with hard examples mined from our real video (RV) dataset, and we observe a significant 3.4\% AP improvement. This improvement demonstrates the significance of mined hard examples  from real videos, which enables the learned model to track through occlusion, large pose deformation, and motion blur. We also show qualitative examples in Fig. \ref{fig:comp-results}.

\subsection{Biased sampling of hard examples} \label{sec:hard-sampling}

We next analyze whether focusing on the hard examples that we mine from unlabeled videos might help to improve the training of our tracker. 
We define a hard example to be a short video clip that includes a broken track which has been identified and rectified by our optimization process. Next, we adopt biased sampling for hard examples during model training, and we list the tracker's performance with different sampling rates in Table \ref{hard-resample}. For every training batch, we sample 50\% examples from RV and the rest from HV.

As shown in Table \ref{hard-resample}, there is a significant increase from introducing biased sampling of hard examples, which clearly illustrates the value of hard example mining. We empirically observe that the tracker achieves the best performance at a hard example sampling rate of around 75\%, which leads to a substantial 3\% AP@0.5 improvement over non-biased sampling. As hard examples are relatively sparse in the dataset, training without biased sampling performs nearly the same as sampling all easy examples.

\subsection{Balancing between hallucinated and real videos}

\begin{table}[t]
\small
\centering
\begin{tabular}{ccc}
\toprule
Video balancing ratio & AP @0.5 & AP @0.75 \\
\midrule
0.00 & 32.0 & 15.4\\
0.25 & 34.2 &	16.4 \\
0.50 & 35.4	& 17.0 \\
0.75 & 35.6	& 17.7 \\
1.00 & 27.4 & 13.1 \\
\bottomrule
\end{tabular}
\caption{\small \textbf{Data balancing ratio.} We compare performance of trackers trained with different balancing ratios between RV and HV data on TAO-Person. A higher ratio indicates more RV data.
}
\label{video-ratio}
\end{table}

In this section, we further investigate the optimal balancing ratio between sampling from HV and RV during training.
We define video balancing ratio as $\frac{|RV|}{|RV| + |HV|}$, in which $|HV|$ and $|RV|$ refers to the quantity of HV and RV in a training epoch respectively. We use the optimal biased sampling rate (i.e. 75\%) during model training.  

As suggested in Table \ref{video-ratio}, there is a significant benefit to adopting a video balancing ratio between 50\% and 75\%. This result indicates the equal importance of both HV and RV. On one hand, HV provides high-quality and high-diversity of pseudo labels for learning robust geometric correspondence of the same instance. On the other hand, RV offers precious hard real videos that enable the tracker to track through dramatic human pose deformation, partial occlusion, etc.  Therefore, it is important to train the tracker on both HV and RV.
While training on either set of data alone does not work as well, training on hard examples from RV alone causes a more significant performance drop, indicating the importance of regularization from HV during model training.

\section{Conclusion}

In this paper, we explored training a deep neural tracker in a weakly supervised setting where we have bounding box annotations and a large pool of unlabeled videos. This is important as manual tracking annotations are extremely expensive to obtain. In detail, we first hallucinated a video sequence by applying recurrent zoom-in/out motion transformations to an image and later injecting various random motion effect transformations to the sequence. Next, we proposed to mine hard training examples from unlabeled videos. We considered hard examples to be video clips where a tracker trained only on hallucinated video would fail. We proposed an optimization-based connecting process to identify and rectify the broken tracklets, generating mined hard examples and their corresponding pseudo labels. Using our hallucinated videos and mined hard videos, we trained a deep neural tracker. The weakly supervised tracker achieved state-of-the-art performance on the challenging MOT17 and TAO-Person datasets, demonstrating the value of our self-generated dataset. On MOT17, we further showed that our self-generated data was complementary to the existing manually-annotated dataset, allowing for further improvements when combined.

{\small
\bibliographystyle{ieee_fullname}
\bibliography{egbib}

\begin{thebibliography}{10}\itemsep=-1pt

\bibitem{andriyenko2011multi}
Anton Andriyenko and Konrad Schindler.
\newblock Multi-target tracking by continuous energy minimization.
\newblock In {\em CVPR 2011}. IEEE, 2011.

\bibitem{berclaz2006robust}
Jerome Berclaz, Francois Fleuret, and Pascal Fua.
\newblock Robust people tracking with global trajectory optimization.
\newblock In {\em 2006 IEEE Computer Society Conference on Computer Vision and
  Pattern Recognition (CVPR'06)}. IEEE, 2006.

\bibitem{berclaz2011multiple}
Jerome Berclaz, Francois Fleuret, Engin Turetken, and Pascal Fua.
\newblock Multiple object tracking using k-shortest paths optimization.
\newblock {\em IEEE transactions on pattern analysis and machine intelligence},
  33(9):1806--1819, 2011.

\bibitem{bergmann2019tracking}
Philipp Bergmann, Tim Meinhardt, and Laura Leal-Taixe.
\newblock Tracking without bells and whistles.
\newblock In {\em Proceedings of the IEEE international conference on computer
  vision}, pages 941--951, 2019.

\bibitem{bewley2016simple}
Alex Bewley, Zongyuan Ge, Lionel Ott, Fabio Ramos, and Ben Upcroft.
\newblock Simple online and realtime tracking.
\newblock In {\em 2016 IEEE International Conference on Image Processing
  (ICIP)}, pages 3464--3468. IEEE, 2016.

\bibitem{braso2020learning}
Guillem Bras{\'o} and Laura Leal-Taix{\'e}.
\newblock Learning a neural solver for multiple object tracking.
\newblock In {\em Proceedings of the IEEE/CVF Conference on Computer Vision and
  Pattern Recognition}, pages 6247--6257, 2020.

\bibitem{Butler:ECCV:2012}
D.~J. Butler, J. Wulff, G.~B. Stanley, and M.~J. Black.
\newblock A naturalistic open source movie for optical flow evaluation.
\newblock In {A. Fitzgibbon et al. (Eds.)}, editor, {\em European Conf. on
  Computer Vision (ECCV)}, Part IV, LNCS 7577, pages 611--625. Springer-Verlag,
  Oct. 2012.

\bibitem{carreira2019short}
Joao Carreira, Eric Noland, Chloe Hillier, and Andrew Zisserman.
\newblock A short note on the kinetics-700 human action dataset.
\newblock {\em arXiv preprint arXiv:1907.06987}, 2019.

\bibitem{dave2020tao}
Achal Dave, Tarasha Khurana, Pavel Tokmakov, Cordelia Schmid, and Deva Ramanan.
\newblock Tao: A large-scale benchmark for tracking any object.
\newblock {\em arXiv preprint arXiv:2005.10356}, 2020.

\bibitem{deng2009imagenet}
Jia Deng, Wei Dong, Richard Socher, Li-Jia Li, Kai Li, and Li Fei-Fei.
\newblock Imagenet: A large-scale hierarchical image database.
\newblock In {\em 2009 IEEE conference on computer vision and pattern
  recognition}, pages 248--255. Ieee, 2009.

\bibitem{gao2018im2flow}
Ruohan Gao, Bo Xiong, and Kristen Grauman.
\newblock Im2flow: Motion hallucination from static images for action
  recognition.
\newblock In {\em Proceedings of the IEEE Conference on Computer Vision and
  Pattern Recognition}, pages 5937--5947, 2018.

\bibitem{he2019revisiting}
Junxian He, Jiatao Gu, Jiajun Shen, and Marc'Aurelio Ranzato.
\newblock Revisiting self-training for neural sequence generation.
\newblock {\em arXiv preprint arXiv:1909.13788}, 2019.

\bibitem{he2017mask}
Kaiming He, Georgia Gkioxari, Piotr Doll{\'a}r, and Ross Girshick.
\newblock Mask r-cnn.
\newblock In {\em Proceedings of the IEEE international conference on computer
  vision}, pages 2961--2969, 2017.

\bibitem{he2016deep}
Kaiming He, Xiangyu Zhang, Shaoqing Ren, and Jian Sun.
\newblock Deep residual learning for image recognition.
\newblock In {\em Proceedings of the IEEE conference on computer vision and
  pattern recognition}, pages 770--778, 2016.

\bibitem{held2016learning}
David Held, Sebastian Thrun, and Silvio Savarese.
\newblock Learning to track at 100 fps with deep regression networks.
\newblock In {\em European Conference on Computer Vision}, pages 749--765.
  Springer, 2016.

\bibitem{IMKDB17}
E. Ilg, N. Mayer, T. Saikia, M. Keuper, A. Dosovitskiy, and T. Brox.
\newblock Flownet 2.0: Evolution of optical flow estimation with deep networks.
\newblock In {\em IEEE Conference on Computer Vision and Pattern Recognition
  (CVPR)}, Jul 2017.

\bibitem{jin2018unsupervised}
SouYoung Jin, Aruni RoyChowdhury, Huaizu Jiang, Ashish Singh, Aditya Prasad,
  Deep Chakraborty, and Erik Learned-Miller.
\newblock Unsupervised hard example mining from videos for improved object
  detection.
\newblock In {\em Proceedings of the European Conference on Computer Vision
  (ECCV)}, pages 307--324, 2018.

\bibitem{kahn2020self}
Jacob Kahn, Ann Lee, and Awni Hannun.
\newblock Self-training for end-to-end speech recognition.
\newblock In {\em ICASSP 2020-2020 IEEE International Conference on Acoustics,
  Speech and Signal Processing (ICASSP)}, pages 7084--7088. IEEE, 2020.

\bibitem{kanazawa2019learning}
Angjoo Kanazawa, Jason~Y Zhang, Panna Felsen, and Jitendra Malik.
\newblock Learning 3d human dynamics from video.
\newblock In {\em Proceedings of the IEEE/CVF Conference on Computer Vision and
  Pattern Recognition}, pages 5614--5623, 2019.

\bibitem{kay2017kinetics}
Will Kay, Joao Carreira, Karen Simonyan, Brian Zhang, Chloe Hillier, Sudheendra
  Vijayanarasimhan, Fabio Viola, Tim Green, Trevor Back, Paul Natsev, et~al.
\newblock The kinetics human action video dataset.
\newblock {\em arXiv preprint arXiv:1705.06950}, 2017.

\bibitem{MOTChallenge2015}
L. Leal-Taix\'{e}, A. Milan, I. Reid, S. Roth, and K. Schindler.
\newblock {MOTC}hallenge 2015: {T}owards a benchmark for multi-target tracking.
\newblock {\em arXiv:1504.01942 [cs]}, Apr. 2015.
\newblock arXiv: 1504.01942.

\bibitem{li2018flow}
Yijun Li, Chen Fang, Jimei Yang, Zhaowen Wang, Xin Lu, and Ming-Hsuan Yang.
\newblock Flow-grounded spatial-temporal video prediction from still images.
\newblock In {\em Proceedings of the European Conference on Computer Vision
  (ECCV)}, pages 600--615, 2018.

\bibitem{li2020improving}
Yandong Li, Di Huang, Danfeng Qin, Liqiang Wang, and Boqing Gong.
\newblock Improving object detection with selective self-supervised
  self-training.
\newblock In {\em European Conference on Computer Vision}, pages 589--607.
  Springer, 2020.

\bibitem{lin2014microsoft}
Tsung-Yi Lin, Michael Maire, Serge Belongie, James Hays, Pietro Perona, Deva
  Ramanan, Piotr Doll{\'a}r, and C~Lawrence Zitnick.
\newblock Microsoft coco: Common objects in context.
\newblock In {\em European conference on computer vision}, pages 740--755.
  Springer, 2014.

\bibitem{meinhardt2021trackformer}
Tim Meinhardt, Alexander Kirillov, Laura Leal-Taixe, and Christoph
  Feichtenhofer.
\newblock Trackformer: Multi-object tracking with transformers.
\newblock {\em arXiv preprint arXiv:2101.02702}, 2021.

\bibitem{Menze2015CVPR}
Moritz Menze and Andreas Geiger.
\newblock Object scene flow for autonomous vehicles.
\newblock In {\em Conference on Computer Vision and Pattern Recognition
  (CVPR)}, 2015.

\bibitem{MOT16}
A. Milan, L. Leal-Taix\'{e}, I. Reid, S. Roth, and K. Schindler.
\newblock {MOT}16: {A} benchmark for multi-object tracking.
\newblock {\em arXiv:1603.00831 [cs]}, Mar. 2016.
\newblock arXiv: 1603.00831.

\bibitem{misra2015watch}
Ishan Misra, Abhinav Shrivastava, and Martial Hebert.
\newblock Watch and learn: Semi-supervised learning for object detectors from
  video.
\newblock In {\em Proceedings of the IEEE Conference on Computer Vision and
  Pattern Recognition}, pages 3593--3602, 2015.

\bibitem{DBLP:conf/icra/OsepVLBL19}
Aljosa Osep, Paul Voigtlaender, Jonathon Luiten, Stefan Breuers, and Bastian
  Leibe.
\newblock Large-scale object mining for object discovery from unlabeled video.
\newblock In {\em International Conference on Robotics and Automation, {ICRA}
  2019, Montreal, QC, Canada, May 20-24, 2019}, pages 5502--5508. {IEEE}, 2019.

\bibitem{porzi2020learning}
Lorenzo Porzi, Markus Hofinger, Idoia Ruiz, Joan Serrat, Samuel~Rota Bulo, and
  Peter Kontschieder.
\newblock Learning multi-object tracking and segmentation from automatic
  annotations.
\newblock In {\em Proceedings of the IEEE/CVF Conference on Computer Vision and
  Pattern Recognition}, pages 6846--6855, 2020.

\bibitem{ren2015faster}
Shaoqing Ren, Kaiming He, Ross Girshick, and Jian Sun.
\newblock Faster r-cnn: Towards real-time object detection with region proposal
  networks.
\newblock In {\em Advances in neural information processing systems}, pages
  91--99, 2015.

\bibitem{roychowdhury2019automatic}
Aruni RoyChowdhury, Prithvijit Chakrabarty, Ashish Singh, SouYoung Jin, Huaizu
  Jiang, Liangliang Cao, and Erik Learned-Miller.
\newblock Automatic adaptation of object detectors to new domains using
  self-training.
\newblock In {\em Proceedings of the IEEE/CVF Conference on Computer Vision and
  Pattern Recognition}, pages 780--790, 2019.

\bibitem{scudder1965probability}
H Scudder.
\newblock Probability of error of some adaptive pattern-recognition machines.
\newblock {\em IEEE Transactions on Information Theory}, 11(3):363--371, 1965.

\bibitem{shao2018crowdhuman}
Shuai Shao, Zijian Zhao, Boxun Li, Tete Xiao, Gang Yu, Xiangyu Zhang, and Jian
  Sun.
\newblock Crowdhuman: A benchmark for detecting human in a crowd.
\newblock {\em arXiv preprint arXiv:1805.00123}, 2018.

\bibitem{shuai2021siammot}
Bing Shuai, Andrew Berneshawi, Xinyu Li, Davide Modolo, and Joseph Tighe.
\newblock Siammot: Siamese multi-object tracking.
\newblock In {\em CVPR}, 2021.

\bibitem{sun2020transtrack}
Peize Sun, Yi Jiang, Rufeng Zhang, Enze Xie, Jinkun Cao, Xinting Hu, Tao Kong,
  Zehuan Yuan, Changhu Wang, and Ping Luo.
\newblock Transtrack: Multiple-object tracking with transformer.
\newblock {\em arXiv preprint arXiv:2012.15460}, 2020.

\bibitem{tang2017multiple}
Siyu Tang, Mykhaylo Andriluka, Bjoern Andres, and Bernt Schiele.
\newblock Multiple people tracking by lifted multicut and person
  re-identification.
\newblock In {\em Proceedings of the IEEE Conference on Computer Vision and
  Pattern Recognition}, pages 3539--3548, 2017.

\bibitem{wang2018deep}
Mei Wang and Weihong Deng.
\newblock Deep visual domain adaptation: A survey.
\newblock {\em Neurocomputing}, 312:135--153, 2018.

\bibitem{wojke2017simple}
Nicolai Wojke, Alex Bewley, and Dietrich Paulus.
\newblock Simple online and realtime tracking with a deep association metric.
\newblock In {\em 2017 IEEE international conference on image processing
  (ICIP)}, pages 3645--3649. IEEE, 2017.

\bibitem{xiang2015learning}
Yu Xiang, Alexandre Alahi, and Silvio Savarese.
\newblock Learning to track: Online multi-object tracking by decision making.
\newblock In {\em Proceedings of the IEEE international conference on computer
  vision}, pages 4705--4713, 2015.

\bibitem{xie2020self}
Qizhe Xie, Minh-Thang Luong, Eduard Hovy, and Quoc~V Le.
\newblock Self-training with noisy student improves imagenet classification.
\newblock In {\em Proceedings of the IEEE/CVF Conference on Computer Vision and
  Pattern Recognition}, pages 10687--10698, 2020.

\bibitem{xu2019spatial}
Jiarui Xu, Yue Cao, Zheng Zhang, and Han Hu.
\newblock Spatial-temporal relation networks for multi-object tracking.
\newblock In {\em Proceedings of the IEEE International Conference on Computer
  Vision}, pages 3988--3998, 2019.

\bibitem{xu2019train}
Yihong Xu, Aljosa Osep, Yutong Ban, Radu Horaud, Laura Leal-Taixe, and Xavier
  Alameda-Pineda.
\newblock How to train your deep multi-object tracker.
\newblock {\em arXiv preprint arXiv:1906.06618}, 2019.

\bibitem{yu2018deep}
Fisher Yu, Dequan Wang, Evan Shelhamer, and Trevor Darrell.
\newblock Deep layer aggregation.
\newblock In {\em Proceedings of the IEEE conference on computer vision and
  pattern recognition}, pages 2403--2412, 2018.

\bibitem{zhang2008global}
Li Zhang, Yuan Li, and Ramakant Nevatia.
\newblock Global data association for multi-object tracking using network
  flows.
\newblock In {\em 2008 IEEE Conference on Computer Vision and Pattern
  Recognition}, pages 1--8. IEEE, 2008.

\bibitem{zhou2020tracking}
Xingyi Zhou, Vladlen Koltun, and Philipp Kr{\"a}henb{\"u}hl.
\newblock Tracking objects as points.
\newblock {\em arXiv preprint arXiv:2004.01177}, 2020.

\bibitem{zhu2020improving}
Yi Zhu, Zhongyue Zhang, Chongruo Wu, Zhi Zhang, Tong He, Hang Zhang, R
  Manmatha, Mu Li, and Alexander Smola.
\newblock Improving semantic segmentation via self-training.
\newblock {\em arXiv preprint arXiv:2004.14960}, 2020.

\bibitem{zoph2020rethinking}
Barret Zoph, Golnaz Ghiasi, Tsung-Yi Lin, Yin Cui, Hanxiao Liu, Ekin~D Cubuk,
  and Quoc~V Le.
\newblock Rethinking pre-training and self-training.
\newblock {\em arXiv preprint arXiv:2006.06882}, 2020.

\bibitem{zou2018unsupervised}
Yang Zou, Zhiding Yu, BVK Kumar, and Jinsong Wang.
\newblock Unsupervised domain adaptation for semantic segmentation via
  class-balanced self-training.
\newblock In {\em Proceedings of the European conference on computer vision
  (ECCV)}, pages 289--305, 2018.

\bibitem{zou2019confidence}
Yang Zou, Zhiding Yu, Xiaofeng Liu, BVK Kumar, and Jinsong Wang.
\newblock Confidence regularized self-training.
\newblock In {\em Proceedings of the IEEE International Conference on Computer
  Vision}, pages 5982--5991, 2019.

\end{thebibliography}
}

\end{document}